\documentclass[11pt,a4paper]{article}
\usepackage[hyperref]{naaclhlt2018}
\usepackage{times}
\usepackage{latexsym}
\usepackage{indentfirst}
\usepackage{amsmath}
\usepackage{amssymb}

\usepackage{url}

\usepackage{amsmath}
\usepackage{amssymb}

\usepackage{graphicx}
\usepackage{subfigure}
\usepackage{booktabs} 
\usepackage{multirow}
\usepackage{natbib}
\usepackage{kotex}

\aclfinalcopy 

\title{GLAC Net: GLocal Attention Cascading Networks for Multi-image Cued Story Generation}

\author{Taehyeong Kim$^1$, Min-Oh Heo$^2$, Seonil Son$^2$, Kyoung-Wha Park$^3$, Byoung-Tak Zhang$^{1,2,3}$  \\
  1 Program in Cognitive Science, 2 Dept. of Computer Sci. and Eng., 3 Program in Neuroscience\\
  Seoul National University \\
  Seoul 08826  South Korea \\
  {\tt  \{thkim,moheo,sison,kwpark,btzhang\}@bi.snu.ac.kr} \\}

\date{}

\begin{document}
\maketitle
\begin{abstract}
The task of multi-image cued story generation, such as visual storytelling dataset (VIST) challenge, is to compose multiple coherent sentences from a given sequence of images. 
The main difficulty is how to generate image-specific sentences within the context of overall images. 
Here we propose a deep learning network model, GLAC Net, that generates visual stories by combining global-local (glocal) attention and context cascading mechanisms. 
The model incorporates two levels of attention, i.e., overall encoding level and image feature level, to construct image-dependent sentences.
While standard attention configuration needs a large number of parameters, the GLAC Net implements them in a very simple way via hard connections from the outputs of encoders or image features onto the sentence generators. 
The coherency of the generated story is further improved by conveying (cascading) the information of the previous sentence to the next sentence serially. 
We evaluate the performance of the GLAC Net on the visual storytelling dataset (VIST) and achieve very competitive results compared to the state-of-the-art techniques. Our code and pre-trained models are available here\footnote{\url{https://github.com/tkim-snu/GLACNet}}.


\end{abstract}

\section{Introduction} 


Deep learning have brought about breakthroughs in processing image, video, speech and audio \cite{lecun2015deep}.
The field of natural language processing has been also interested in deep learning, e.g., sentence classification \cite{kim2014convolutional,iyyer2015deep}, language modeling \cite{bengio2003neural,mikolov2013distributed}, machine translation \cite{sutskever2014sequence,Bah2014nmt,wu2016google}, and question answering \cite{hermann2015teaching}.
Naturally, bridging images and texts by deep learning has been following \cite{belz20018from} such as image captioning \cite{vinyals2015show,xu2015showattend,Karpathy2017dva}, visual question answering \cite{antol2015vqa,kim2016multimodal}, and image generation from caption \cite{reed2016generative,zhang2017stackgan}.


The task of multi-image cued story generation is one of interesting visual-linguistic challenges to generate story of multiple coherent sentences from a given sequence of images.
The main difficulty is how to generate image-specific sentences within the context of overall images. 
Additionally, it is harder than object recognition or image captioning since it needs fine-grained object recognition and context understanding in the images.
Recently, visual storytelling dataset (VIST) was released for the task of multi-image cued story generation, which is composed of five-sentence stories, descriptions and the corresponding sequences of five images \cite{huang2016visual}.

Here we propose a deep learning network model that generates visual stories by combining global-local (glocal) attention and context cascading mechanisms. 
To focus on the image-specific appropriateness of them, we develop two levels of attention, i.e., overall encoding level (global) and image feature level (local).
In the image sequence encoders, the global context of the storyline is encoded using bi-directional LSTMs on features of five images, we give attention on the context (\textit{global} attention).
Additionally, we give \textit{local} attention to image features directly. 
Then both of them are combined and sent to RNN-based sentence generators.
While standard attention configuration needs a large number of parameters, we implement them in a very simple way via hard connections from the outputs of encoders or image features onto the sentence generators.
To improve further the coherency of the generated stories, we design to convey the last hidden vector in the sentence generator to the next sentence generator as an initial hidden vector.


%
This paper is organized as follows. 
Section 2 presents related works to positioning. 
In Section 3 we show briefly dataset, section 4 explains the proposed models.
Section 5 shows their experimental results.
Finally, section 6 draws the conclusion.

\section{Related Work}
\textbf{Text Comprehension}  As similar works without visual cues, there are text comprehension tasks such as bAbI tasks \cite{weston2015babi}, SQuAD \cite{pranav2016squad} and Story Cloze Test \cite{nasrin2016clozetest}.
They have been widely used to benchmark new algorithms on document comprehension or story understanding.

\textbf{Visually-grounded Comprehension}  Since AlexNet \cite{kriz2012imgnet} as a milestone,  object recognition/detection methods have grown explosively and outperformed human ability to capture objects in accuracy aspect \cite{geirhos2017comphuman}. 
As used in our model, those visual features turned out to be also available as general features for scene description \cite{karpathy2016densecap,Karpathy2017dva}, image captioning with attention \cite{xu2015showattend}, and image/video question answering about the stories \cite{MovieQA,kim2017deepstory}.

\textbf{Story Generation from Images}
In the first work for image cued sentence generation \cite{farhadi2010every}, the triplet - $<$object, action, scene$>$ was predicted for an input image using MRF, and used for searching or generating with templates.
In the deep learning era, \citet{jain2017story} utilized the VIST dataset to translate description-to-story without images.
\citet{Liu2017let-photos-talk} developed semantic embedding of the image features on the bi-directional recurrent architecture to generate a relevant story to the pictures.


\section{VIST Dataset}
\begin{figure}[h!]
\includegraphics[width=\columnwidth]{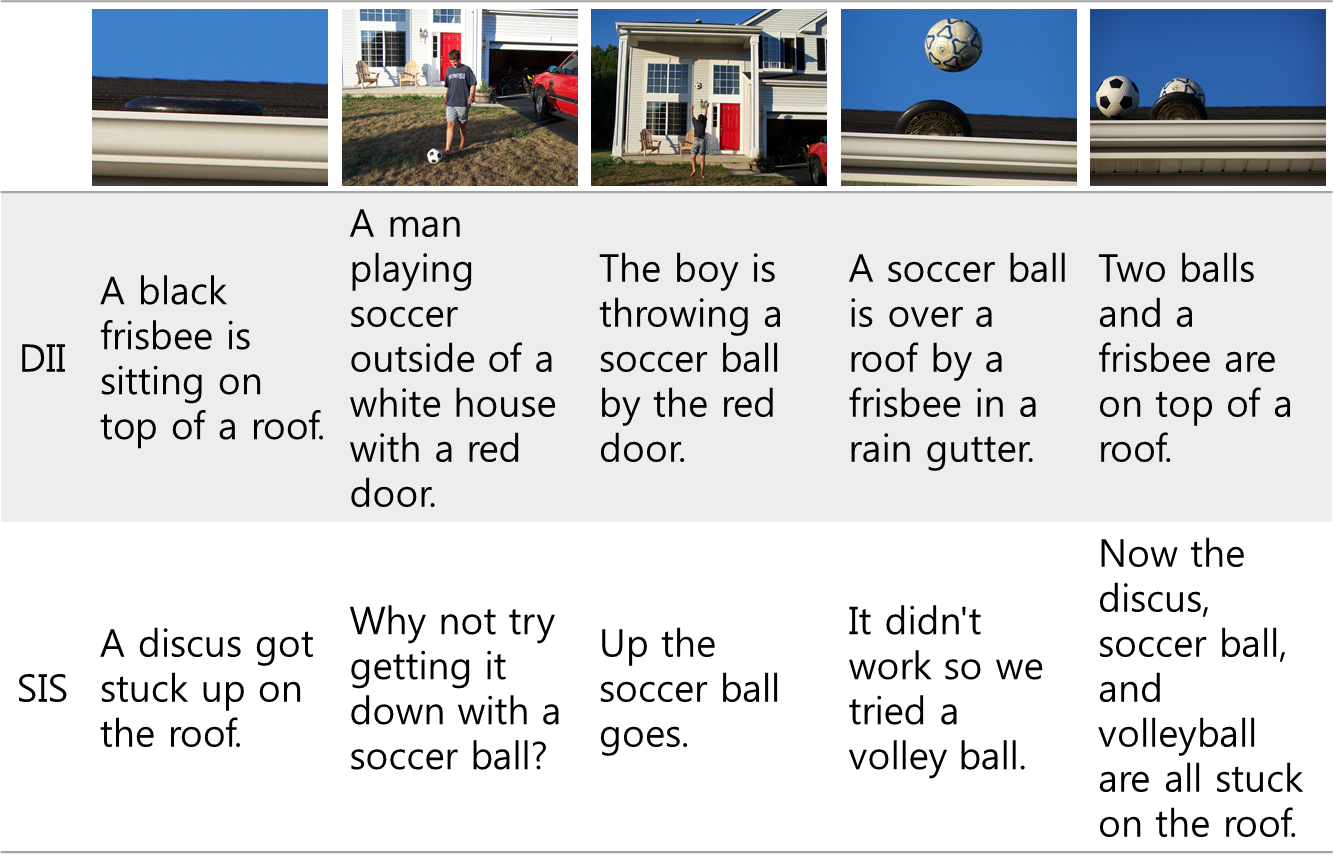}
\caption{A VIST dataset example. DII: Descriptions of images in isolation. SIS: Stories of images in sequence.}
\label{fig:vist}
\end{figure}
VIST dataset is a dataset of story-like image sequences paired with: (1) descriptions for each image in isolation (DII) ($\sim80\%$ only), and (2) descriptions to form a narrative over an image sequence (images/sentences aligned each) (SIS) as shown in Figure \ref{fig:vist}.
It consists of 50,200 sequences (stories) using 209,651 images (train: 40,155, validation: 4,990, test: 5,055).

\section{Approaches to Story Generation}

The main difficulty of the multi-image cued story generation is to keep the overall context of the story while generating a well-aligned sentence for each image. 
To tackle it, we introduce 2 key ideas: (1) utilizing two-level attention mechanism in the encoder part, and (2) conveying the hidden state to the next sentence generator.

\subsection{Two-levels of Attention}

Soft attention mechanism \cite{Bah2014nmt} utilizes additional weights on the inter-related outputs of the nodes, which improves the performance of the basic encoder-decoder model in machine translation.
In the task of story generation from image sequences, however, each sentence should be visually grounded on not only each image but also overall context.
To represent these relationship, we design to deliver the two-channel information to each decoder (1) from low-level image features, and (2) from high-level encoded features together.
We implement them via a simple hard attention mechanism that focuses sequentially on the encoder output when generating story-like sentences.

\begin{figure*}[h!] \centering
\includegraphics[width=\linewidth]{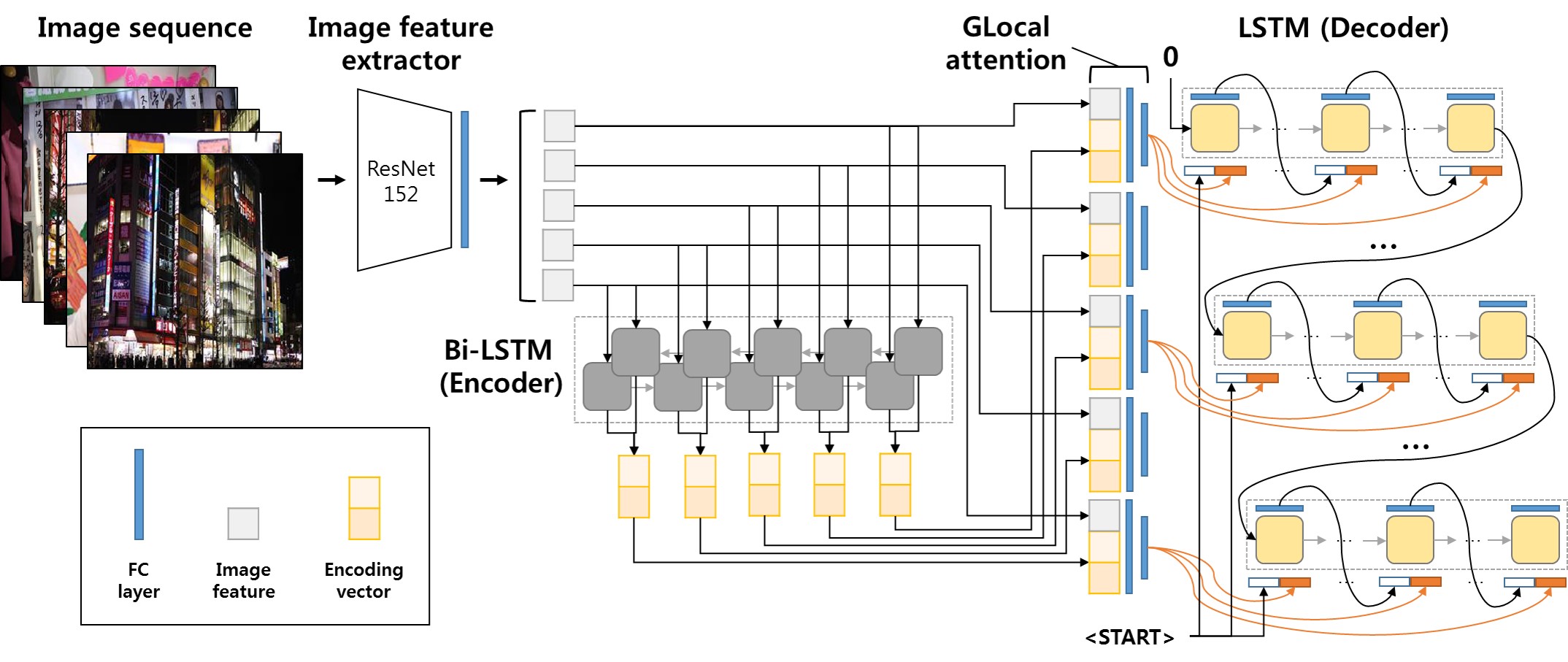}
\caption{The global-local attention cascading (GLAC) network model for visual story generation. Note: activation function (ReLU), dropout, batch normalization, and softmax layer are omitted for readability.}
\label{fig:model}
\end{figure*}

In the sequence to sequence configuration, we can choose one of the followings as encoder of image sequences: concatenation of image features (for short sequences with the same length), uni-directional RNN, and bi-directional RNN.
We choose bi-RNN because it is better for aggregated representations. 
For image specificity, we feed both bi-RNN outputs and image-specific features to decoders.
The outputs of bi-RNN include overall information of the sequence (global).
On the other hand, image-specific features are constrained only on the image (local).
The glocal vectors are obtained by concatenating image-specific features and bi-RNN outputs.
Including glocal vector as decoder inputs can be seen as an 'hard' attention mechanism, which emphasizes both image-specific and overall information.

\subsection{Model Details}

Figure \ref{fig:model} illustrates deep learning architecture for the story generation proposed in this paper. 
(1) the features of each image are extracted using the ResNet-152 \cite{he2015resnet}. 
(2) The extracted features are sequentially fed into the bi-LSTM so that the context of the images can be evenly reflected in the entire story.
The glocal vectors made up of bi-LSTM outputs and image-specific features go through the fully connected layers. After that, it is concatenated to the word tokens in order to be used as inputs to the decoder.
Note that one glocal vector is used until the decoder meets an '$<$END$>$' token which denotes the end of sentence;
five glocal vectors for each image works the same as described above.
The cascading mechanism conveys the hidden state (context) of the previous sentence to the next sentence. 
The hidden state of the LSTM is initialized to zeros only at the beginning of the first sentence of the story for maintaining story context.

As a simple heuristics to avoid duplicates in the resulting sentence, we sample words one hundred times from the word probability distribution of the LSTM output, and choose the most frequent word from the sampled pool.
This reduces the number of repetitive expressions and improve the diversity of the generated sentences.
On the process of generating sentences of the story, We also count the selected words.
The selection probabilities of the words are decreased according to the frequency of each word as Equation 1, and normalized.
\begin{equation}
\hat{p}(word)=p(word) \times \frac{1}{1 + k \cdot count_{word}}
\end{equation}
where \textit{k} is a constant for sensitivity.

To build grammatically correct sentences, the probabilities of some function words such as prepositions and pronouns are not changed regardless of the frequency of occurrence.

\subsection{Network Training}
The training images are resized to $256 \times 256$ before training, and then augmented with random cropping of $224 \times 224$ accompanied by a horizontal flip at the training time. 
Pixels are normalized to [0,1]. 
Learning rate and weight decay are set to 0.001 and 1e-5 respectively and optimized with Adam optimizer. 
Each word is embedded into a vector of 256 dimensions, and the LSTM is trained with teacher forcing manner. 
Also, we apply batch normalization and dropout layers to prevent overfitting and improve the performance. 
Batch size are set to 64, and the training data is reshuffled at every epoch.

\section{Experiments and Discussion}
\subsection{Experiment Settings}

In order to evaluate the effects of the GLAC Net, we performed an ablation study as follows. 
The first model is a simple LSTM Seq2Seq network.
In the second model, we remove context cascading from the full GLAC Net architecture. 
The third and fourth models are for testing the effect of global and local attention respectively.
In the fifth, we remove the post processing routines to avoid word duplication when generating sentences.
The last is the complete GLAC Net model.
\subsection{Results and Discussion}

We evaluate trained models on the VIST dataset. 
The evaluation criteria are perplexity and METEOR metric and the results are shown in Table \ref{tab:result}. 
Compared with the performance of baselines \cite{huang2016visual}, the GLAC Net is also competitive without beam search methods. 
From the results of 'GLAC Net (-Count)' and 'Baselines (-Dups)' in Table \ref{tab:result}, the heuristics are helpful to reduce redundant sentences and improve the METEOR score.
Compared to LSTM Seq2Seq models, GLAC Net-based model shows better performance in general.
Although the differences are not much significant between the GLAC Net experiment settings, the complete GLAC Net shows the best overall performance.

\begin{figure}[!h]
\includegraphics[width=\columnwidth]{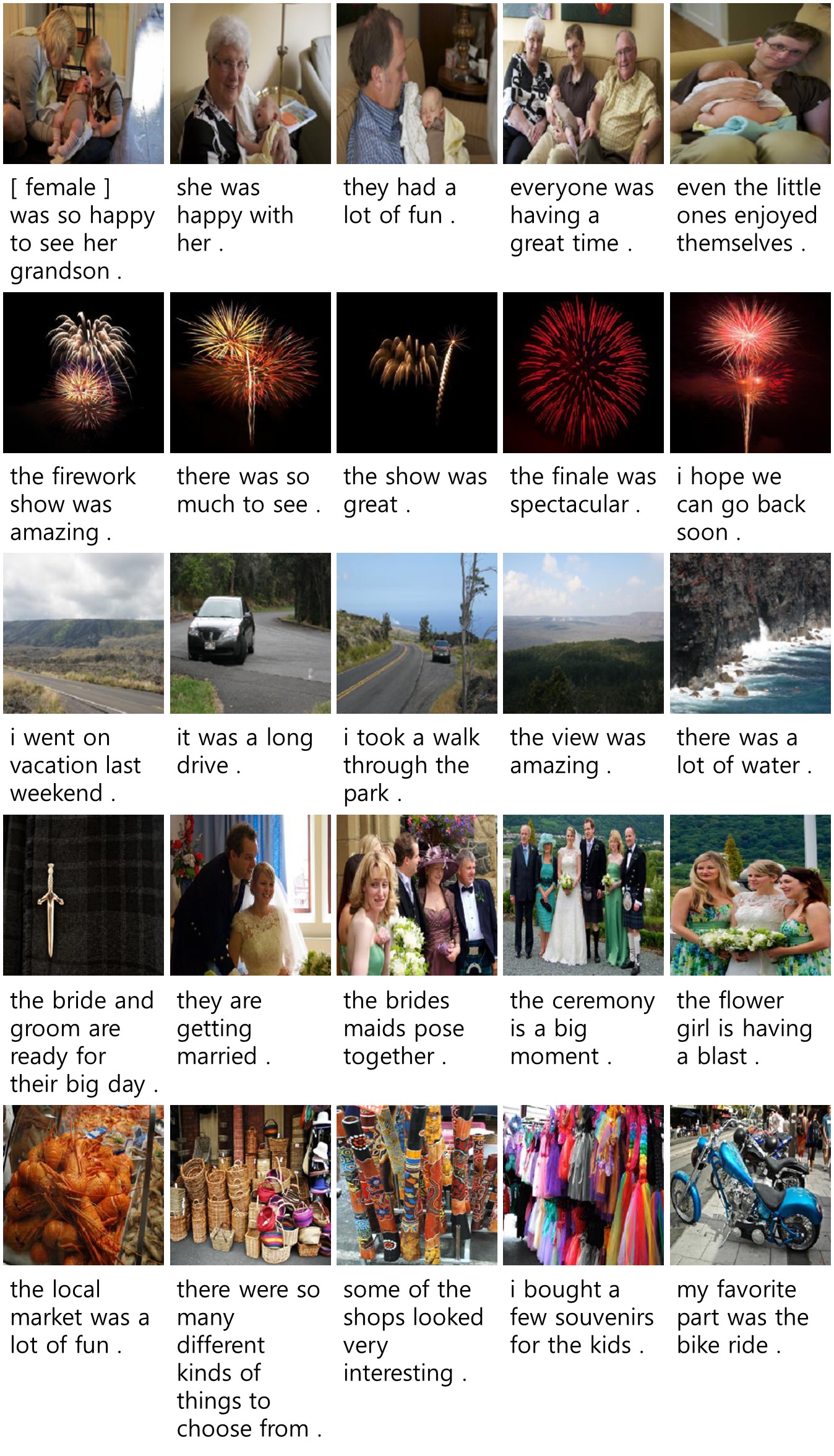}
\caption{Samples of multi-image cued story generation results}
\label{fig:result}
\end{figure}

We also consider human evaluation criteria to choose better models \cite{humanEvalVist}.
Figure \ref{fig:result} shows the examples of the generated stories with the test dataset. 
The context of successive images is well reflected, and the content of each image is properly described. 
However, the story development is slightly monotonous. 
Sometimes the content is different from the specific image or the sentence is a little awkward. 
It is observed that most generated sentences have simple structures.

\begin{table}
  \begin{center}
    \resizebox{\columnwidth}{!}{
    \begin{tabular}{l|c|c|c} 
      \textbf{Used} & \textbf{Valid } & \textbf{Test} & \textbf{METEOR} \\
      \textbf{Attention} & \textbf{Perplexity} & \textbf{Perplexity} & \textbf{Score} \\
      \hline
      \hline
      Baselines (Beam=10) & - & - & 0.2313 \\
      Baselines (Greedy) & - & - & 0.2776 \\
      Baselines (-Dups) & - & - & 0.3011 \\
      Baselines (+Grounded) & - & - & 0.3142 \\
      \hline
      LSTM Seq2Seq & 21.89 & 22.18 & 0.2721 \\
      GLAC Net (-Cascading) & 20.24 & 20.54 & \textbf{0.3063} \\
      GLAC Net (-Global) & 18.32 & 18.47 & 0.2913 \\
      GLAC Net (-Local) & 18.21 & 18.33 & 0.2996 \\
      GLAC Net (-Count) & 18.13 & 18.28 & 0.2823 \\
      GLAC Net & \textbf{18.13} & \textbf{18.28} & 0.3014 \\
    \end{tabular}
    }
    \caption{Results from experiment settings. Baselines are reported in \cite{huang2016visual}.}
    \label{tab:result}
  \end{center}
\end{table}

\section{Conclusion and Future Work}

We proposed the GLAC Net that uses glocal attention and context cascading mechanisms to generate stories from a sequence of images. 
The model is designed to maintain the overall context of the story from the image sequence and to generate context-aware sentences for each image. 
In the experiment using the VIST dataset, the proposed model is proved to be effective and competitive in storytelling task according to the crowd-sourced human evaluation results with METEOR score $\sim$0.3.

Although the experimental results are promising, visual storytelling task is remaining as a challenge.
We are planning to extend and refine the GLAC architecture to further improvement of its performance.
Also, generating various stories based on the purpose and theme from the same image sequence would be the following topic to be explored in the future works.


\section*{Acknowledgments}

This work was partly supported by the Korean government (R0126-16-1072-SW.StarLab, 2017-0-01772-VTT , 2018-0-00622-RMI, 10060086-RISF).



\bibliography{naaclhlt2018}
\bibliographystyle{acl_natbib}

\end{document}